\pdfoutput=1

\documentclass[11pt]{article}
\usepackage{authblk}
\usepackage{graphicx}
\usepackage[dvipsnames]{xcolor}
\usepackage{acl}

\usepackage{times}
\usepackage{latexsym}
\usepackage{amsmath}
\usepackage{subcaption}
\usepackage[export]{adjustbox}

\usepackage{pifont}
\newcommand{\cmark}{\ding{51}}
\newcommand{\xmark}{\ding{55}}

\usepackage[T1]{fontenc}

\usepackage[utf8]{inputenc}

\usepackage{microtype}

\title{Detection, Disambiguation, Re-ranking: \\ Autoregressive Entity Linking as a Multi-Task Problem}

\author[1]{\bf Khalil Mrini}
\author[2]{\bf Shaoliang Nie}
\author[2]{\bf Jiatao Gu}
\author[2]{\bf Sinong Wang}
\author[2]{\\ \bf Maziar Sanjabi}
\author[2]{\bf Hamed Firooz}
\affil[1]{University of California, San Diego, La Jolla, CA 92093 \protect\\ \small{\texttt{khalil@ucsd.edu}}}
\affil[2]{Meta AI, Menlo Park, CA 94025 \protect\\ \small{\texttt{\{snie, jgu, swang, maziars, hfirooz\}@fb.com}}}

\definecolor{specialyellow}{HTML}{BF9000}
\definecolor{specialblue}{HTML}{4a86e8}
\definecolor{specialgreen}{HTML}{38761d}
\definecolor{specialred}{HTML}{cc0000}

\begin{document}
\maketitle

\begin{abstract}
We propose an autoregressive entity linking model, that is trained with two auxiliary tasks, and learns to re-rank generated samples at inference time. Our proposed novelties address two weaknesses in the literature. First, a recent method proposes to learn mention detection and then entity candidate selection, but relies on predefined sets of candidates. We use encoder-decoder autoregressive entity linking in order to bypass this need, and propose to train mention detection as an auxiliary task instead. Second, previous work suggests that re-ranking could help correct prediction errors. We add a new, auxiliary task, match prediction, to learn re-ranking. Without the use of a knowledge base or candidate sets, our model sets a new state of the art in two benchmark datasets of entity linking: COMETA in the biomedical domain, and AIDA-CoNLL in the news domain. We show through ablation studies that each of the two auxiliary tasks increases performance, and that re-ranking is an important factor to the increase. Finally, our low-resource experimental results suggest that performance on the main task benefits from the knowledge learned by the auxiliary tasks, and not just from the additional training data. 
\end{abstract}

\section{Introduction}

Entity linking \cite{zhang2010entity, han2011collective} is the task of linking entity mentions in a text document to concepts in a knowledge base. It is a basic building block used in many NLP applications, such as question answering \cite{yu2017improved, dubey2018earl, shah2019kvqa}
, word sense disambiguation \cite{raganato2017word, uslu2018fastsense}, text classification \cite{basile2015deep, scharpf2021towards}, and social media analysis \cite{liu2013entity, yamada2015end}.

\begin{figure}
    \centering
    \includegraphics[width=\columnwidth]{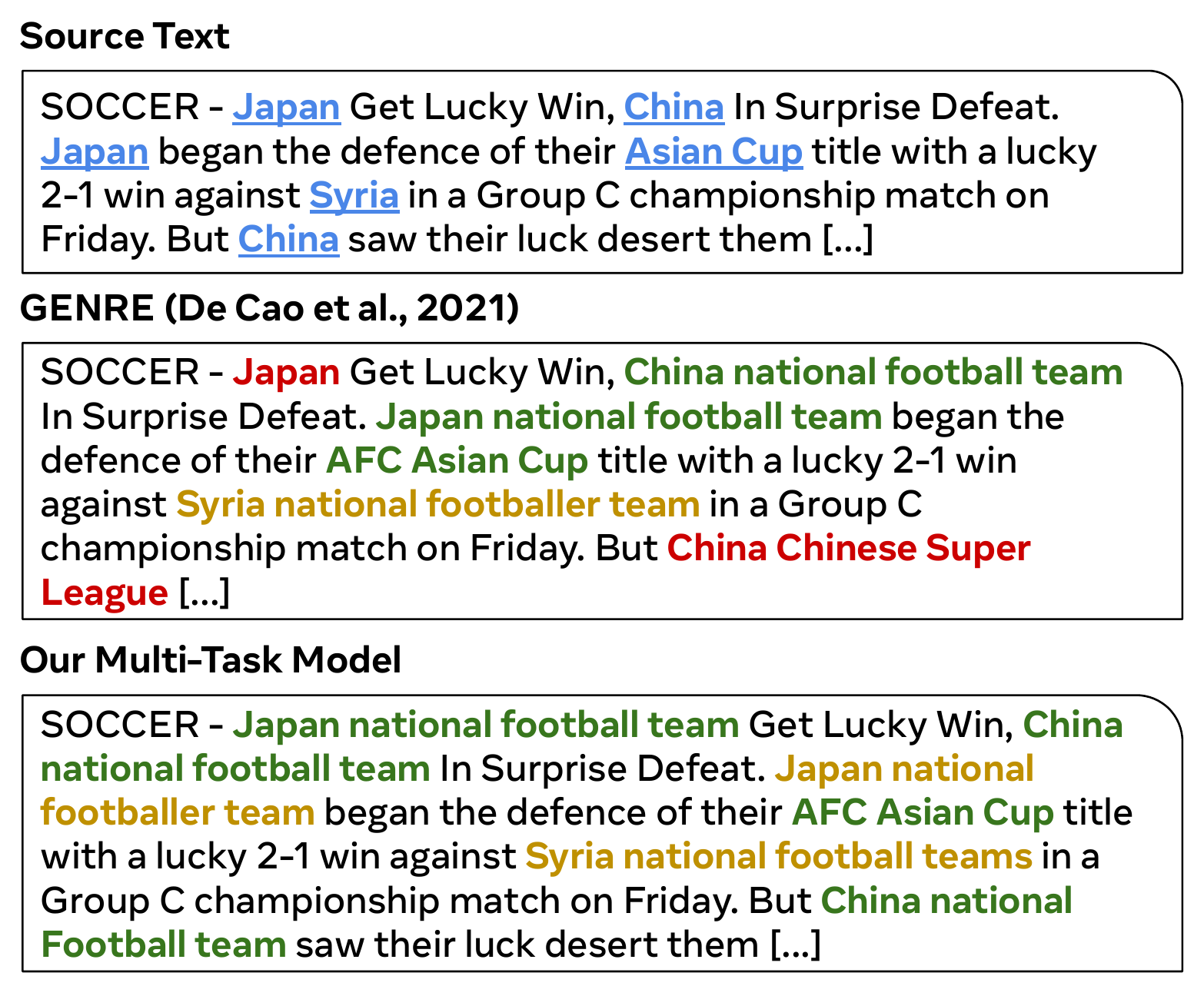}
    \caption{Example of an Entity Linking (EL) source text and generated outputs. Entity mentions to be recognized and disambiguated are denoted in {\color{specialblue}\underline{\textbf{{blue}}}} in the source text. In the outputs, \textbf{\color{specialred}{red}} denotes errors, \textbf{\color{specialgreen}{green}} denotes correct answers, \textbf{\color{specialyellow}{yellow}} denotes close matches.}
    \label{fig1}
\end{figure}

Early definitions decompose the task of entity linking (EL) into two subtasks: Mention Detection (MD) and Entity Disambiguation (ED). Many statistical and LSTM-based methods propose to cast EL as a two-step problem, and optimize for both MD and ED \cite{guo2013link, luo2015joint, cornolti2016piggyback, ganea2017deep}. 

Recent entity linking methods based on language models propose to cast entity linking as a single, end-to-end trained task \cite{broscheit2019investigating, poerner2020bert, el2020correlation}, rather than a two-subtask problem. An example is autoregressive entity linking \cite{petroni2021kilt, decao2020autoregressive}, which formulates entity linking as a language generation problem using an encoder-decoder model. A more recent approach \cite{de2021highly} increases performance, and is instead based on a two-step architecture: first mention detection with a transformer encoder, and then autoregressive candidate selection with an LSTM. However, this candidate selection module needs a predefined set of candidate mentions.

Methods based on word embedding models \cite{basaldella2020cometa} propose to learn entity disambiguation by mapping embedding spaces. Their high accuracy at 10 results show that re-ranking could increase entity linking performance. 

\textbf{Contributions.} In this paper, we propose an autoregressive entity linking method, that is trained jointly with two auxiliary tasks, and learns to re-rank generated samples at inference time. Our proposed novelties address two weaknesses in the literature.

First, instead of the two-step method \cite{de2021highly} that learns to detect mentions and then to select the best entity candidate from a predefined set, we propose to add mention detection as an \textit{auxiliary} task to encoder-decoder-based autoregressive EL. By using encoder-decoder-based autoregressive EL, we bypass the need for a predefined set of candidate mentions, while preserving the benefit of the knowledge learned from mention detection for the main EL task.

Second, previous work suggests that re-ranking could correct prediction errors \cite{basaldella2020cometa}. We propose to train a second, new auxiliary task, called \textit{Match Prediction}. This task teaches the model to re-rank generated samples at inference time. We define match prediction as a classification task where the goal is to identify whether entities in a first sentence were correctly disambiguated in the second sentence. We train this second task with samples generated by the model at each training epoch. At inference time, we then rank the generated samples using our match prediction scores.

Our multi-task learning model outperforms the state of the art in two benchmark datasets of entity linking across two domains: COMETA \cite{basaldella2020cometa} from the biomedical and social media domain, and AIDA-CoNLL \cite{hoffart2011robust} from the news domain. We show through three ablation study experiments that each auxiliary task provides improvements on the main task. Then, we show that using our model's match prediction module to re-rank generated samples at inference time plays an important role in increasing performance. Finally, we devise three experiments where we train auxiliary tasks with a smaller dataset. Results suggest that our model's performance is not only due to more training datapoints, but also due to our auxiliary task definition.

\section{Related Work}

\textbf{Entity Linking (EL).} Entity Linking is often \cite{hoffart2011robust, steinmetz2013semantic, piccinno2014tagme, de2021highly} trained as two tasks: Mention Detection (MD) and Entity Disambiguation (ED). Mention detection is the task of detecting entity mention spans, such that an entity mention $m$ is represented by start and end positions. A mention $m$ refers to a concept in a given knowledge base. Entity disambiguation is the task of finding the right knowledge base concept for an entity mention, thereby \textit{disambiguating} its meaning.

Early EL methods \cite{hoffart2011robust, steinmetz2013semantic, daiber2013improving} rely on probabilistic approaches. \citet{hoffart2011robust} propose a probabilistic framework for MD and ED, based on textual similarity and corpus occurrence. They test their framework using the entity candidate sets available in the AIDA-CoNLL dataset.

More recently, neural methods propose to train end-to-end EL models. \citet{francis2016capturing} propose a convolutional neural EL model to take into account windows of context.

\citet{kolitsas2018end} propose a neural model for joint mention detection and entity disambiguation. They use a bidirectional LSTM \cite{hochreiter1997long} to encode spans of entities. They then embed candidate entities and train layers to score the likelihood of a match.

\citet{sil2018neural} introduce an LSTM-based model that uses multilingual embeddings for zero-shot transfer from English-language knowledge bases.

\textbf{EL as Language Modeling.} Language modeling approaches have enabled new, end-to-end definitions of the entity linking task. These new settings enable to bypass the two-step MD-then-ED setting for entity linking, and propose to cast entity linking as a single task.

\citet{broscheit2019investigating} propose to reformulate end-to-end EL problem as a token-wise classification over the entire set of the vocabulary. Their model is based on BERT \cite{devlin2019bert}. The training combines mention detection, candidate generation, and entity disambiguation. If an entity is not detected, then the prediction is \textit{O}. If an entity is detected, the classification head has to classify it as the corresponding particular entity within the vocabulary.

\citet{decao2020autoregressive} propose an autoregressive setting for EL. They use BART \cite{lewis2020bart} and cast entity linking as a language generation task. In this setting, the input is the source sentence with the entity mention. The goal is to generate an annotated version of the input sentence, such that the entity mention is highlighted and mapped to a knowledge base concept. Brackets and parentheses are used to annotate the entity mention and concept: ``\textit{I took the [flu shot] (influenza vaccine).}''. They then introduce a constrained beam search to force the model to annotate. \citet{de2021multilingual} is a multilingual extension of this work.

\textbf{EL as Embedding Space Mapping.} Language models like BERT, as well as embedding models like FastText \cite{bojanowski2017enriching}, enable to retrieve context-aware representations of entities and knowledge base concepts.

\citet{basaldella2020cometa} propose to map the embeddings of entity mentions to the embeddings of knowledge base concepts. 
They find that the right mapping is more often found among the ten closest concept embeddings (accuracy at 10) rather than being the closest concept embedding (accuracy at 1). Their results suggest that generated sample re-ranking could improve entity linking systems. 

Concurrently, \citet{wu2020scalable} propose a method that uses re-ranking for zero-shot retrieval of entities. They use entity definition embeddings to find candidate entities from a knowledge base, and then train a cross-encoder to re-rank the candidates.

\citet{basaldella2020cometa} also introduce the COMETA dataset: an entity linking benchmark based on social media user utterances on medical topics, and linked to the SNOMED-CT biomedical knowledge base \cite{donnelly2006snomed}. The dataset has four splits, based on whether the dev/test set entities are seen during training (stratified) or not (zeroshot), and on whether the entity mapping is context-specific (specific) or not (general). \citet{liu2021self} propose a self-alignment pre-training scheme for entity embeddings, and show that it benefits the context-free splits (stratified general and zeroshot general). \citet{liu2021fast} propose MirrorBERT: a data-augmented approach for masked language models. \citet{lai2021bert} and \citet{kong-etal-2021-zero} propose convolution-based and graph-based methods, respectively, for embedding mapping between entities and knowledge base concepts.

All of the above methods use knowledge base concepts. In our biomedical entity linking setting, we choose the harder zeroshot specific split. We propose to use the language modeling task setting instead of the embedding mapping method. We therefore bypass the need to embed each and every knowledge base concept, whereas only a small portion (<10\%) of the SNOMED-CT knowledge base concepts are used in the COMETA dataset.

\section{Multi-Task Learning for Autoregressive Entity Linking}

We propose an autoregressive entity linking model, that is trained along with two auxiliary tasks, and uses re-ranking at inference time.

\begin{figure*}
    \centering
    \includegraphics[width=\textwidth]{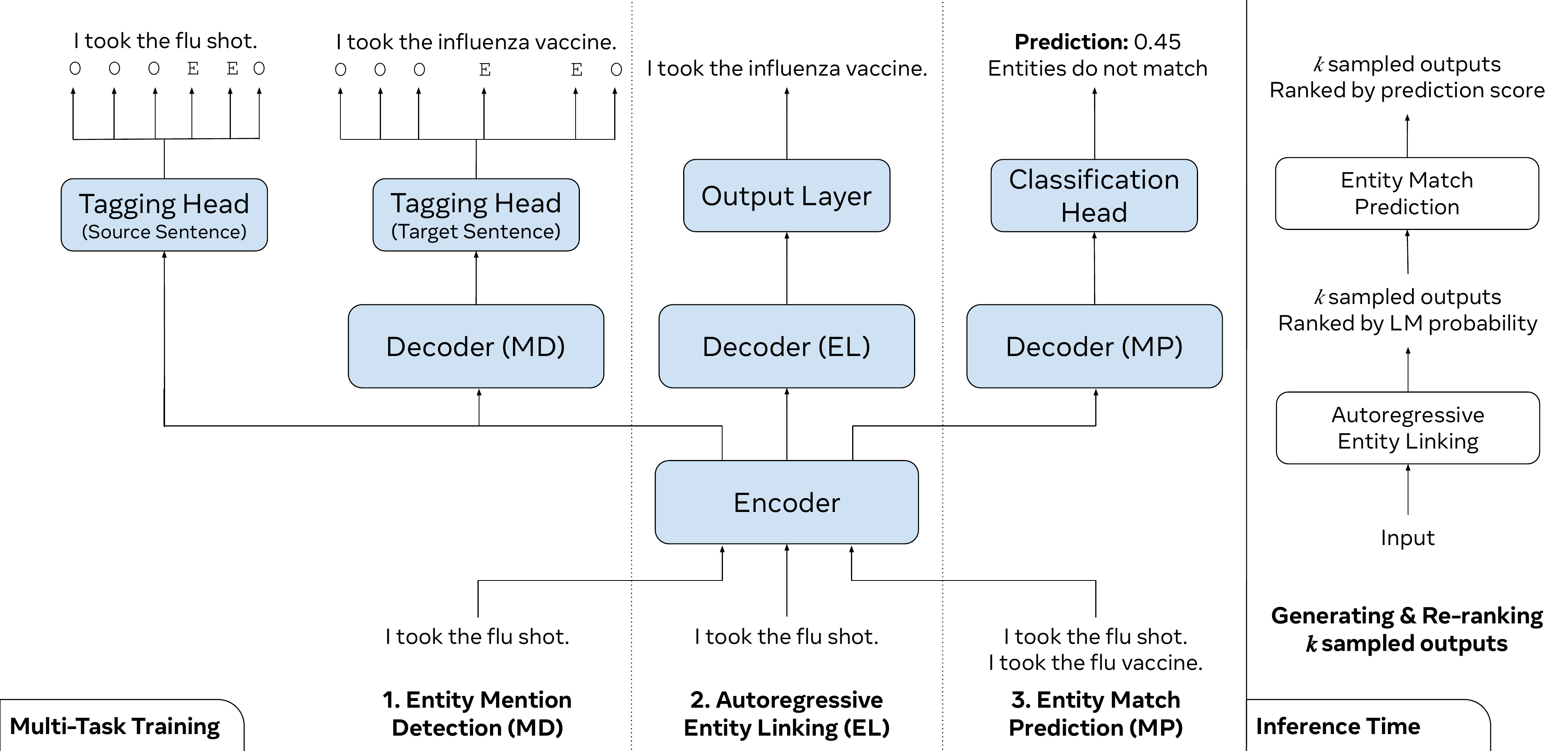}
    \caption{Architecture of our proposed multi-task autoregressive entity linking model. Each task is trained using a shared encoder and a task-specific decoder and output layer. The auxiliary mention detection task uses datasets derived from one entity linking dataset, whereas the match prediction task uses sampled outputs. At inference time, we use the match prediction module to re-rank generated samples.
    }
    \label{fig2}
\end{figure*}

In this section, we first describe the main entity linking task. Then, we define the two auxiliary tasks: Mention Detection and a new task, called \textit{Match Prediction}. Third, we train our multi-task learning architecture with a weighted objective. Finally, we propose to use the match prediction module for re-ranking during inference. An overview of our architecture is in Figure \ref{fig2}. 

\subsection{Autoregressive Entity Linking}

We train autoregressive entity linking as a language generation task. We follow the setting of the encoder-decoder model of \citet{decao2020autoregressive}. They train their model to generate the input sentence containing both the entity mention \textit{and} the target entity, annotated with parentheses and brackets. For simplicity, we omit these annotations from the examples in the figures. 

For entity linking (EL), we optimize the following negative log-likelihood loss:

\begin{equation}
    \mathcal{L}_\mathrm{EL} = - \sum_{i=1}^{N}\mathrm{log}P(y_i|y_1, ..., y_{i-1},\mathbf{x})
    \label{eq1}
\end{equation}

\noindent where $\mathbf{x}$ is the input sentence, and $\mathbf{y}$ is the output sentence of length $N$.

\subsection{Entity Mention Detection}

We introduce mention detection (MD) as an auxiliary task to encoder-decoder autoregressive EL, in order for the knowledge learned from MD to benefit the main EL task, while bypassing the need for predefined candidate sets. 
MD teaches the model to distinguish tokens that are part of entities from tokens that are not part of any entity. As a result, this task is in essence a token-wise binary classification task. 
\citet{broscheit2019investigating} propose a similar task definition, but combine entity detection with entity disambiguation. Their task definition is a classification task over the entire knowledge base vocabulary, rather than our binary setting.

In this task, we train the model to predict where the tokens of the entities are in the input sentence and in the target (annotated) sentence. Therefore, this auxiliary task has to output two sequences of entity indicators: ``\textit{E}'' for entity mention or concept tokens, and ``\textit{O}'' for all other tokens. To train our model to generate sequences for the input and target sentences, we augment our existing dataset. We create two datasets of the same size: the first has sequences of entity indicators for the input sentences, and the second has sequences of entity indicators for the target sentences.

As shown at the left of Figure \ref{fig2}, we use two different tagging heads for mention detection: one for the input sentence, and one for the output sentence. We use two tagging heads as the model learns different mappings from two different kinds of input. For the input sentence, we feed the encoder embeddings to the first tagging head. We cast this as a classification problem. For mention detection on the output sentence, we use a separate decoder, and feed this decoder's embeddings to the second tagging head. We cast this task as a generation task. For both tasks, we optimize a cross entropy (CE) loss. In summary, we optimize the following loss function for mention detection (MD):

\begin{equation}
    \begin{split}
    \mathcal{L}_\mathrm{MD} = & \mathrm{CE}\left(Enc(\mathbf{x}), Ent(\mathbf{x})\right) \\ &+ \mathrm{CE}\left(Dec(Enc(\mathbf{x})), Ent(\mathbf{y})\right)
    \end{split}
    \label{eq2}
\end{equation}

\noindent where $Enc(\cdot)$ is the encoder representation, $Dec(\cdot)$ is the decoder representation, and $Ent(\cdot)$ indicates the corresponding sequence of entity indicators.

The method of \citet{de2021highly} has two steps, where the first step is to detect mentions. Here, mention detection is an auxiliary task rather than a main part of the pipeline. We employ encoder-decoder autoregressive EL as our main end-to-end pipeline.

\subsection{Entity Match Prediction}

In their biomedical entity linking experiments using word embedding space mapping, \citet{basaldella2020cometa} find that accuracy at 10 is often more than double the accuracy at 1. They then suggest that re-ranking could significantly improve performance. We build on this observation to introduce the second auxiliary task: entity match prediction (MP). The goal of this task is to teach the model to re-rank generated samples based on the input sentence, with the aim to help narrow the gap with the accuracy at 10 scores.

The input to this task is composed of two sentences: the first one is the input sentence, and the second is a sentence where entity mentions are replaced by entities that may or may not be the matching target entities. We train the model to predict whether the entities match (score of 1) or not (score of 0) between both sentences. The entity match must be complete -- all target entities must be generated -- for a score of 1.

At regular intervals during training, we generate $k$ samples for each input sentence using beam search on the autoregressive entity linking part of the trained model. We then form $k$ sentence pairs. The corresponding ground truth label for a given sentence pair indicates whether the entities match or not. This data generation setting exposes the model to its own successes and failures in the main entity linking task.

It may be the case that no generated sample contains entities that match the input sentence, and therefore that all labels for a pair are 0. In this case, the model would not be shown what an example of matching entities looks like. To mitigate this issue, we decide to add one additional sentence pair, where the second sentence is the target sentence used in the autoregressive entity linking training. We add this additional sentence pair to all datapoints for consistency.

We train entity match prediction using a mean squared error loss:

\begin{equation}
    \begin{split}
        \mathcal{L}_\mathrm{MP} = & \left(P^{\mathrm{MP}}(\mathbf{\hat{y}}|\mathbf{x}) - 1 \right)^2 \\ & + \sum_{i=1}^{k} \left(P^{\mathrm{MP}}(\mathbf{y}^s_i|\mathbf{x}) - \hat{y}^{\mathrm{MP}}_i \right)^2
    \end{split}
    \label{eq3}
\end{equation}

\noindent where $\mathbf{\hat{y}}$ is the target sentence, $\mathbf{y}^s_i$ is the $i$-th generated sample, $P^{\mathrm{MP}}(\cdot|\cdot)$ is the probability that the entities in the left-hand sequence match the ones in the right-hand sequence, and $\hat{y}^{\mathrm{MP}}_i$ is the ground truth label for entity match prediction for the $i$-th generated sample.

\citet{de2021highly} propose to rank candidate concepts from a predefined set after the detecting entity mentions. In our case, we do not learn to rank predefined sets of candidates, nor do we rank concepts. Instead, we generate sentences using beam search, and propose to learn to re-rank them.

\subsection{Multi-Task Learning}

We propose to optimize simultaneously for all three tasks using a single loss function. We set one weight for each auxiliary task. We discuss the task weight hyperparameter tuning in \S \ref{weight}.

Given the losses defined in equations \ref{eq1}, \ref{eq2}, and \ref{eq3}, our loss function for multi-task learning is as follows:

\begin{equation}
    \mathcal{L}_\mathrm{MTL} = \mathcal{L}_\mathrm{EL} + \lambda_\mathrm{MD} \mathcal{L}_\mathrm{MD} + \lambda_\mathrm{MP} \mathcal{L}_\mathrm{MP}
\end{equation}

\noindent where $\lambda_\mathrm{MD}$ and $\lambda_\mathrm{MP}$ are the auxiliary task weights for mention detection and match prediction, respectively.

As shown in Figure \ref{fig2}, we use three separate decoders for training: one for each task. We use two separate tagging heads for mention detection. For the match prediction task, we feed the last decoder output to the classification head. This follows the training scheme of BART \cite{lewis2020bart} for sentence classification tasks.

Our proposed multi-task definition is inspired by our prior work \cite{mrini-etal-2021-joint, mrini-etal-2021-gradually}. In our prior research papers, we introduce multi-task learning architectures for biomedical question summarization and entailment. We find that closely related tasks benefit each other during learning, through either multi-task learning or transfer learning \cite{mrini-etal-2021-ucsd}.

Our model architecture is also inspired by MT-DNN \cite{liu2019multi}, a multi-task model that obtained state-of-the-art results across many NLP tasks involving sentence representation. In the MT-DNN architecture, the encoder is shared across tasks, and prediction heads are task-specific. Nonetheless, other multi-task architectures remain compatible with our auxiliary tasks and re-ranking, which are the novelties we focus on in this work.

\subsection{Inference-time Re-ranking}

In order to bridge some of the gap between accuracy at 1 and accuracy at 10 \cite{basaldella2020cometa}, we propose to use the entity match prediction module to re-rank generated samples. The right side of Figure \ref{fig2} illustrates the process.

At inference time, we first generate $k$ samples ranked by their language modeling probability. We then use the separate entity match prediction (MP) decoder to predict an entity match probability. To do so, we input the source sentence and a generated sample to the MP decoder. We use the resulting MP probabilities to re-rank the $k$ generated samples. We select the sample with the highest MP probability to compute the evaluation metrics.

\section{Experiments}

\subsection{Datasets and Setup}

\begin{table}[]
    \centering
    \begin{tabular}{l r r r}
        \hline
         & \multicolumn{2}{c}{\bf AIDA-CoNLL} & \bf COMETA \\
        \bf Split & \bf Documents & \bf Mentions & \bf Mentions \\ \hline
        Train & 942 & 18,540 & 13,714 \\
        Dev & 216 & 4,791 & 2,018 \\
        Test & 230 & 4,485 & 4,283 \\ \hline
    \end{tabular}
    \caption{Statistics of Entity Linking benchmark datasets.}
    \label{tab:data}
\end{table}

We use two benchmark datasets for English-language entity linking. We use the standard data splits for both datasets, as detailed in Table \ref{tab:data}.

\textbf{AIDA-CoNLL} \cite{hoffart2011robust} is a dataset consisting of annotated news articles from the Reuters Corpus \cite{lewis2004rcv1}. The knowledge base concepts come from the titles of the English-language Wikipedia. Each news article contains multiple entity mentions. Articles are sometimes too long for the maximum sequence length of our model. We follow \citet{de2021highly} and cut the articles into separate chunks. We use the Micro-F1 metric for evaluation. We only evaluate mentions present in the knowledge base, following the \textit{In-KB} setting \cite{roder2018gerbil}, in line with previous work \cite{decao2020autoregressive, de2021highly}. 
This dataset contains candidates for each entity mention. We do not use entity candidates, although several baselines do \cite{kolitsas2018end, martins2019joint, de2021highly}.

\textbf{COMETA} \cite{basaldella2020cometa} is a dataset of biomedical entity mentions from social media (Reddit) utterances. In this dataset, each user-written utterance contains exactly one entity mention. 
The metric used to evaluate this dataset is accuracy at 1 (Acc@1). We measure Acc@1 by checking whether the correct knowledge base concept is present in the top generated sample.
We use the zeroshot specific split, where the entity mention and disambiguation pairs in the test set are not seen during training, and the entity linking is context-specific.

\begin{figure}
    \centering
    \begin{subfigure}[b]{\columnwidth}
        \centering
        \includegraphics[width=\textwidth, frame]{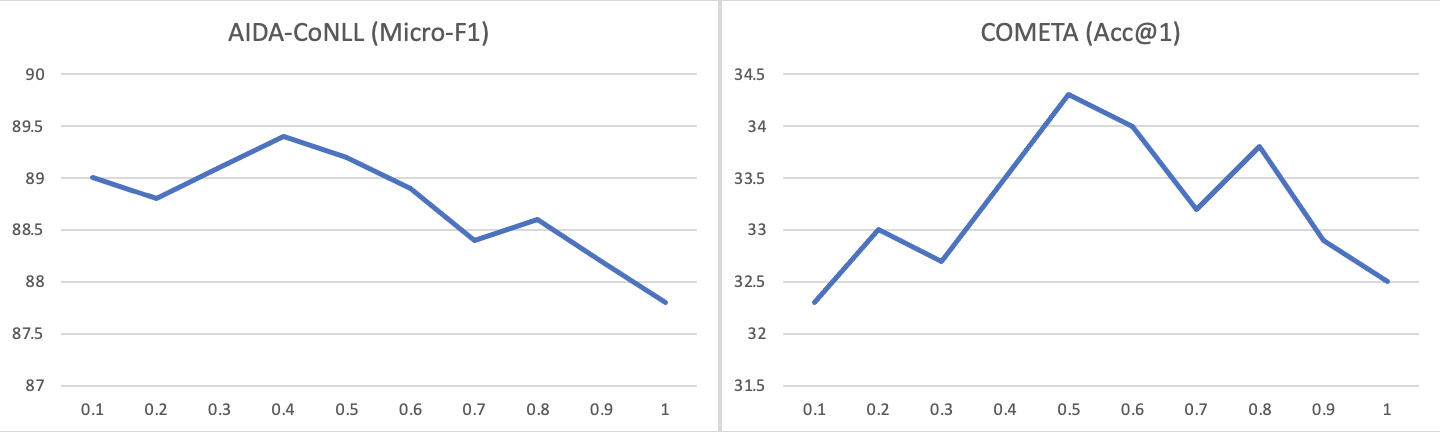}
        \caption{Choosing the optimal $\lambda_{\mathrm{MD}}$, setting $\lambda_{\mathrm{MP}} = 0.3$.}
    \end{subfigure}
    ~
    \begin{subfigure}[b]{\columnwidth}
        \centering
        \includegraphics[width=\textwidth, frame]{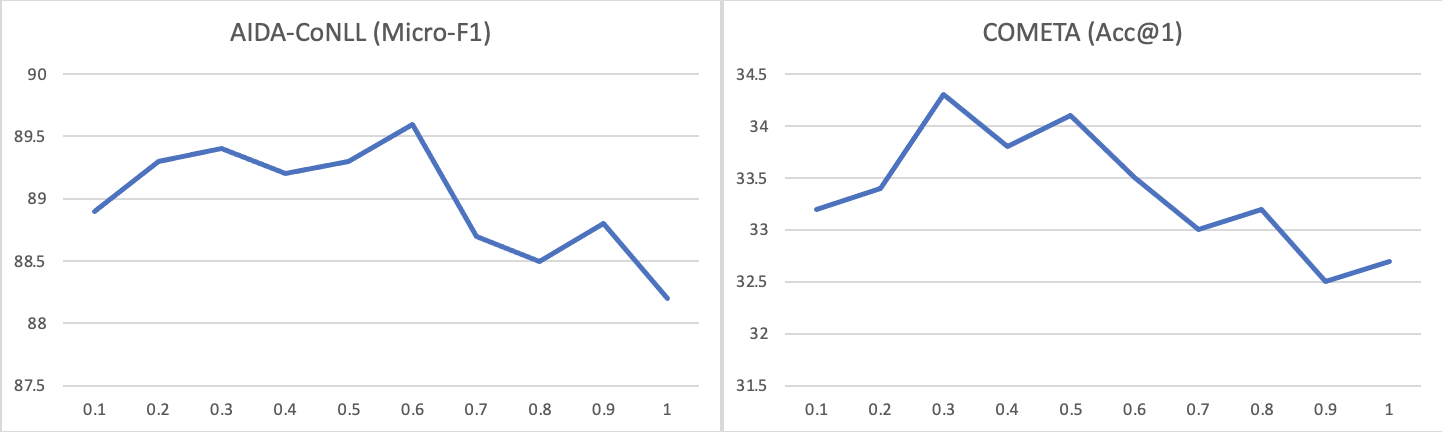}
        \caption{Choosing the optimal $\lambda_{\mathrm{MP}}$, given the optimal $\lambda_{\mathrm{MD}}$.}
    \end{subfigure}
    \caption{Task weight tuning on the dev set for Mention Detection (MD) and Match Prediction (MP). We first optimize for $\lambda_{\mathrm{MD}}$ (a), and then $\lambda_{\mathrm{MP}}$ (b).}
    \label{tuning}
\end{figure}

\subsection{Training Details}

We use BART Large \cite{lewis2020bart} as our base model. We use three decoders, all initialized from the same checkpoint decoder. We found in initial experiments that separate decoders for all tasks benefit the main EL task. We train for 100 epochs on AIDA-CoNLL, and for 10 epochs on COMETA.

\subsection{Task Weight Tuning}
\label{weight}

For each dataset, we optimize the auxiliary task weights $\lambda_{\mathrm{MD}}$ for mention detection, and $\lambda_{\mathrm{MP}}$ for match prediction. We select these hyperparameters based on the highest performance in Micro-F1 (AIDA-CoNLL) or accuracy at 1 (COMETA) on the dev set.

We trial all values from $0.1$ to $1.0$ with $0.1$ increments, for both task weights. We start by optimizing $\lambda_{\mathrm{MD}}$ given $\lambda_{\mathrm{MP}} = 0.3$, and then optimize $\lambda_{\mathrm{MP}}$ given the optimal $\lambda_{\mathrm{MD}}$ weights. The results are in Figure \ref{tuning}. The graphs show that performance on the main entity linking task can vary visibly when the weights of the auxiliary tasks change. The variation is likely due to the large auxiliary task datasets, which could dominate training. Moreover, the optimal task weights are different for every dataset and domain: we find that the optimal auxiliary task weights are $\lambda_{\mathrm{MD}} = 0.4$ and $\lambda_{\mathrm{MP}} = 0.6$ for AIDA-CoNLL, and $\lambda_{\mathrm{MD}} = 0.5$ and $\lambda_{\mathrm{MP}} = 0.3$ for COMETA. We use these task weights for the next experiments.

\subsection{Results and Discussion}

\textbf{AIDA-CoNLL.} The test results for the AIDA-CoNLL dataset are on Table \ref{aida}. Our model establishes a new state of the art for this task.

Compared to the state-of-the-art encoder-decoder autoregressive EL model on AIDA-CoNLL~\cite{decao2020autoregressive}, our method shows a 2.0-point improvement in Micro-F1 score. 
This increase shows that our model is able to correct some errors with the re-ranking at inference time, and that our multi-task setting benefits the main entity linking task.

Our model scores a Micro-F1 0.2 higher than the model of \citet{de2021highly}. However, \citet{de2021highly} use a predefined candidate set of concepts, whereas the encoder-decoder autoregressive EL models -- including our own -- do not. This shows that our model is able to bypass the knowledge base, and that our method leverages language modeling to gain knowledge of the news domain.

\begin{table}[]
    \centering
    \begin{tabular}{l r}
        \hline
        \bf Method & \bf Micro-F1 \\ \hline
        \citet{hoffart2011robust} & 72.8 \\
        \citet{steinmetz2013semantic} & 42.3 \\
        \citet{daiber2013improving} & 57.8 \\
        \citet{10.1162/tacl_a_00179} & 48.5 \\
        \citet{piccinno2014tagme} & 73.0 \\
        \citet{kolitsas2018end} & 82.4 \\
        \citet{peters-etal-2019-knowledge} & 73.7 \\
        \citet{broscheit2019investigating} & 79.3 \\
        \citet{martins2019joint} & 81.9 \\
        \citet{van2020rel} & 80.5 \\
        \citet{fevry-etal-2020-entities} & 76.7 \\
        \citet{kannan-ravi-etal-2021-cholan} & 83.1 \\
        \citet{de2021highly} & 85.5 \\ \hline
        \multicolumn{2}{l}{\bf Encoder-Decoder Autoregressive EL Models} \\
        \citet{decao2020autoregressive} & 83.7 \\
        Our model & \bf 85.7 \\ \hline
    \end{tabular}
    \caption{Results on the AIDA-CoNLL test set.}
    \label{aida}
\end{table}

\textbf{COMETA.} There are no predefined sets of candidate concepts in the COMETA dataset. In this task, there is a knowledge base of biomedical concepts from which the model can choose. Similarly to our AIDA-CoNLL setting, our model does not use the knowledge base.

We consider three baselines for our biomedical entity linking benchmark. The first baseline is the embedding mapping method of \citet{basaldella2020cometa}. They use BioBERT and a max-margin loss with negative target embeddings. The second baseline is the BERT- and classification-based method of \citet{broscheit2019investigating}. We train this baseline by classifying tokens into the concepts present in the COMETA dataset, as opposed to the entire vocabulary of $350K$ knowledge base concepts. This is for computational purposes, as a $350K$-way classification would be difficult to train. The third baseline is the autoregressive, single-task model of \citet{decao2020autoregressive}. We train this baseline as a reference point for our model. We do not include \citet{de2021highly} as a baseline, as their method uses predefined sets of candidate concepts, and COMETA does not include them.

The test results of the COMETA dataset experiments are on Table \ref{cometa}. Our model is able to exceed over five percentage points the baselines that use the knowledge base concepts. This shows that our method can efficiently generalize without the need for a knowledge base, but only through learning about the biomedical domain. Note that we use the zeroshot specific split here, where the entity mention and disambiguation pairs in the test set are not seen during training. Moreover, our model exceeds the autoregressive single-task baseline by 1.5\%. This increase shows that our multi-task setting and re-ranking can generalize, and increase performance under zeroshot settings.

\begin{table}[]
    \centering
    \begin{tabular}{l r}
        \hline
        \bf Method & \bf Acc@1 \\ \hline
        \citet{basaldella2020cometa} & 27.0 \\
        \citet{broscheit2019investigating} & 24.5 \\ \hline
        \multicolumn{2}{l}{\bf Encoder-Decoder Autoregressive EL Models} \\
        \citet{decao2020autoregressive} & 30.9 \\
        Our model & \bf 32.4 \\ \hline
    \end{tabular}
    \caption{Results on the COMETA test set.}
    \label{cometa}
\end{table}

\subsection{Ablation Studies}

We perform two types of ablation studies to analyze the added value of our novelties. First, we evaluate how do the two auxiliary tasks and the re-ranking impact entity linking performance. Second, we implement a low-resource scenario for the auxiliary tasks, as we ask whether the main task benefits more from the knowledge learned the auxiliary tasks, or from the additional training data.

\begin{table}[]
    \centering
    \begin{tabular}{c c c c c}
        \hline
         & & & \bf \small AIDA-CoNLL & \bf \small COMETA \\
        \bf MD & \bf MP & \bf Rk & \bf Micro-F1 & \bf Acc@1 \\ \hline
        \multicolumn{5}{l}{\bf Ablation of Auxiliary Tasks and Re-ranking} \\
        \xmark & \xmark & \xmark & 83.7 & 30.9 \\ \hline
        \multicolumn{5}{l}{\bf Ablation of Auxiliary Tasks} \\
        \cmark & \xmark & \xmark & 84.3 & 31.2 \\
        \xmark & \cmark & \cmark & 85.4 & 32.1 \\ \hline
        \multicolumn{5}{l}{\bf Ablation of Re-ranking} \\
        \cmark & \cmark & \xmark & 84.8 & 31.5 \\ \hline
        \multicolumn{5}{l}{\bf MD, MP and Re-ranking (Ours)} \\
        \cmark & \cmark & \cmark & 85.7 & 32.4 \\ \hline
    \end{tabular}
    \caption{Results of the ablation studies on the test sets. We perform ablation studies on Mention Detection (\textbf{MD}), Match Prediction (\textbf{MP}), and the re-ranking of generated samples (\textbf{Rk}).}
    \label{ablations}
\end{table}

\textbf{Auxiliary Tasks and Re-ranking.}
Our main novelties are multi-task learning with two auxiliary tasks, and the re-ranking of generated samples at inference time. The first auxiliary task, mention detection, aims to preserve the knowledge learned from detecting mentions of entities, while allowing the encoder-decoder model to bypass the need for predefined sets of entity candidates. The second auxiliary task, match prediction, aims to teach the model how to predict whether entities were correctly disambiguated given an input sentence and a generated sample.

We perform ablation studies to gauge the added value of each task and re-ranking. We perform three additional experiments, keeping the same number of model parameters. First, we remove the match prediction training objective ($\lambda_{\mathrm{MP}} = 0.0$), and therefore also remove the re-ranking, but we keep the optimally weighted mention detection objective.  Second, we remove the mention detection training objective by setting $\lambda_{\mathrm{MD}} = 0.0$, but we keep the optimally weighted mention prediction objective, along with the re-ranking. Third, we keep both optimally weighted auxiliary tasks, but remove the inference-time re-ranking of generated samples. Finally, we compare our results to \citet{decao2020autoregressive} as it does not have both auxiliary tasks nor the re-ranking.

We show the results of all ablation experiments on the dev sets in Table \ref{ablations}. The lowest scores are obtained when both auxiliary tasks and re-ranking are ablated. This shows the added value of all of our main novelties on the main entity linking task. In addition, each auxiliary task individually increases performance, as shown on the second and third row of results. The auxiliary match prediction task along with re-ranking provide a larger performance increase than the auxiliary mention detection task alone. This could be due to the fact that the match prediction task gets a larger number of samples to train on. Finally, the difference in performance between our model and the re-ranking ablation study shows that re-ranking of generated samples is an important contribution to the final performance. This result backs the suggestion of \citet{basaldella2020cometa} that re-ranking can bridge some of the gap between Acc@1 and Acc@10.

\textbf{Impact of additional training data.}
In this subsection, we ask whether the main task benefits more from the knowledge learned by the auxiliary tasks, or from the large sizes of the auxiliary task datasets. The mention detection task has two datapoints for every EL datapoint, while the match prediction task has $k + 1 = 11$ datapoints for every EL datapoint. Therefore, in a given training epoch, there are more datapoints to train on for the auxiliary tasks in comparison with the main task.

We devise three experiments to gauge whether a lower amount of training datapoints for auxiliary tasks impacts the main task results. We propose a low-resource regimen of training for auxiliary tasks, such that we bring the ratio of training datapoints down to 1:1 between the auxiliary tasks and the main task. We train on one out of every two MD datapoints, and on one of out every 11 MP datapoints. In other words, we skip 50\% of the training data of the MD task, and 91\% of the training data of the MP task. We spread out the input such that, at each training step, the model sees one EL input sentence, one MD input sentence, and one MP input sentence pair. In each epoch, we skip the same datapoints so that the model only sees a reduced number of training datapoints.

In the first experiment, we train for both auxiliary tasks on a train set ratio of 1:1 with the main task. In the second and third experiments, we apply the low-resource setting only to the mention detection task, and only to the match prediction task, respectively. In all three experiments, we keep the same selection of skipped datapoints for each task, and we keep re-ranking.

\begin{table}[]
    \centering
    \begin{tabular}{c c c c}
        \hline
        \multicolumn{2}{c}{\bf \small \% of Train Set} & \bf \small AIDA-CoNLL & \bf \small COMETA \\
        \bf MD & \bf MP & \bf Micro-F1 & \bf Acc@1 \\ \hline
        \multicolumn{4}{l}{\bf Ablation of Auxiliary Tasks and Re-ranking} \\
        0\% & 0\% & 83.7 & 30.9 \\
        \hline
        \multicolumn{4}{l}{\bf Low-Resource Experiments} \\
        50\% & 9\% & 84.5 & 32.0 \\
        50\% & 100\% & 85.4 & 31.4  \\
        100\% & 9\% & 84.5 & 31.8 \\ \hline
        \multicolumn{4}{l}{\bf No Low-Resource (Ours)} \\
        100\% & 100\% & 85.7 & 32.4 \\ \hline
    \end{tabular}
    \caption{Results on the test sets of the low-resource experiments. We reduce the training datasets of the auxiliary mention detection \textbf{MD} and match prediction \textbf{MP} tasks to measure the benefit of multi-task learning. }
    \label{lowres}
\end{table}

We show the results of the low-resource experiments in Table \ref{lowres}. For reference, we add the results from our model and the model without auxiliary task nor re-ranking of \citet{decao2020autoregressive}. The results show that globally, there is a slight decrease in performance when the training set is smaller, compared to our model. However, the low-resource experiments show a significant increase in performance compared to the ablation experiment of the first row. This shows that our proposed method's edge does not only come from the additional training data, but also from our formulation of the auxiliary tasks, and the re-ranking of generated samples.

\section{Conclusions}

We propose a multi-task learning and re-ranking approach to autoregressive entity linking. Our main two novelties address two weaknesses in the literature. First, whereas the two-step method of \citet{de2021highly} improves performance, it relies on predefined sets of entity candidates. We propose to instead train mention detection as an auxiliary task to autoregressive EL, in order to bypass the need for entity candidate sets, and to preserve the knowledge learned by mention detection. Second, previous work suggests that a sizeable portion of errors could be corrected with re-ranking. We propose to use samples generated at training time to teach the model to re-rank outputs.

Our model establishes a new state of the art in both COMETA and AIDA-CoNLL. The increases in performance across both datasets show that our model can learn and leverage domain-specific knowledge, without using a candidate set or a knowledge base. To analyse our model, we devise three ablation study experiments, and show that our model benefits from both auxiliary tasks and re-ranking. In particular, we show that re-ranking plays a major role in increasing entity linking scores. Then, we propose three low-resource experiments for auxiliary tasks. The results show that our model's performance is not only due to additional training datapoints, but also due to how we defined our auxiliary tasks. 

\section*{Acknowledgements}

KM performed this work during an internship at Meta AI (previously Facebook AI). We thank Jingbo Shang and Ndapa Nakashole for insightful discussions. We thank the anonymous reviewers for their feedback.

\section*{Ethical Considerations}

This work deals with user-generated text in the medical domain. However, our work and models should not be used as text understanding tools for real-life medical systems without human supervision and verification. Our system is not error-free, and using it could lead to a misunderstanding of the true intentions of people seeking medical care.


\bibliography{anthology,custom}
\bibliographystyle{acl_natbib}

\clearpage

\appendix

\end{document}